\documentclass[11pt, a4paper]{article}
\usepackage{fullpage}
\usepackage{amsfonts}
\usepackage{graphicx}
\usepackage{amsmath}
\usepackage{indentfirst}
\usepackage{times}
\begin{document}

\newpage
\begin{center}
\begin{LARGE}
\textbf{Natural Language Processing for Information Extraction}
\end{LARGE}
\end{center}
\begin{center}
Sonit Singh
\end{center}
\begin{center}
Department of Computing, Faculty of Science and Engineering, Macquarie University, Australia
\end{center}
\begin{LARGE}
\textbf{Abstract}
\end{LARGE} 
\vspace{0.2cm}

With rise of digital age, there is an explosion of information in the form of news, articles, social media, and so on. Much of this data lies in unstructured form and manually managing and effectively making use of it is tedious, boring and labor intensive. This explosion of information and need for more sophisticated and efficient  information handling tools gives rise to Information Extraction(IE) and Information Retrieval(IR) technology. Information Extraction systems takes natural language text as input and produces structured information specified by certain criteria, that is relevant to a particular application. Various sub-tasks of IE such as Named Entity Recognition, Coreference Resolution, Named Entity Linking, Relation Extraction, Knowledge Base reasoning forms the building blocks of various high end Natural Language Processing (NLP) tasks such as Machine Translation, Question-Answering System, Natural Language Understanding, Text Summarization and Digital Assistants like Siri, Cortana and Google Now. This paper introduces Information Extraction technology, its various sub-tasks, highlights state-of-the-art research in various IE subtasks, current challenges and future research directions.
 
\section{Introduction to Information Extraction}
Data is now a kind of capital, on par with financial and human capital in creating new digital products and services. With the explosion of information in the form of news, corporate files, medical records, government documents, court hearing and social media, everyone is flooded with information overload. Most of this information is unstructured i.e. free text and thus makes hard to have reasoning and interpretation. Natural Language Processing (NLP) refers to the use of computational methods to process spoken or written form of such free text which acts as a mode of communication commonly used by humans(Assal et al., 2011). There are lot many processes involved in the pipeline of NLP. At the syntactic level, statements are segmented into words, punctuation (i.e. tokens) and each token is assigned with its label in the form of noun, verb, adjective, adverb and so on (Part of Speech Tagging). At the semantic level, each word is analyzed to get the meaningful representation of the sentence. Hence, the basic task of NLP is to process the unstructured text and to produce a representation of its meaning. The higher level tasks in NLP are Machine Translation (MT), Information Extraction (IE), Information Retrieval (IR), Automatic Text Summarization (ATS), Question-Answering System, Parsing, Sentiment Analysis, Natural Language Understanding (NLU) and Natural Language Generation (NLG). Information Extraction (IE) refers to the use of computational methods to identify relevant pieces of information in document generated for human use and convert this information into a representation suitable for computer based storage, processing, and retrieval (Wimalasuriya and Dua, 2010). The input to IE system is a collection of documents (email, web pages, news groups, news articles, business reports, research papers, blogs, resumes, proposals, and so on) and output is a representation of the relevant information from the source document according to some specific criteria. The ability of human beings to effectively make use of this vast amount of information is low as this task is quite boring, tedious and consume lot of time. This explosion of information and need for more sophisticated and efficient information handling tools highlighted the need of information extraction and retrieval technology (Neil et al., 1998).
Information Extraction technologies helps to efficiently and effectively analyze free text and to discover valuable and relevant knowledge from it in the form of structured information. Hence, the goal of IE is to extract salient facts about pre-specified types of events, entities, or relationships, in order to build more meaningful, rich representations of their semantic content, which can be used to populate databases that provide more structured input. 

\section{Information Extraction Tasks}

Information Extraction is often an early stage in pipeline for various high level tasks such as Question Answering Systems (Molla et al., 2006), Machine Translation, event extraction, user profile extraction , and so on. IE is very useful for many commercial applications such as Business Intelligence, automatic annotation of web pages, text mining, and knowledge management (Muludi et al., 2011). Various sub-tasks involved in IE are: Named Entity Recognition (NER), Named Entity Linking (NEL), Coreference Resolution (CR), Temporal Information Extraction, Relation Extraction (RE), Knowledge Base Construction and Reasoning. Shared Tasks in these various sub-tasks of IE have significantly contributed to their progress and provides the fair evaluation of state of the art models. These Shared Tasks include bio-information retrieval (bio-IR), bio-Named Entity Recognition (bio-NER), bio-Information Extraction (bio-IE), bio-Natural Language Processing (bio-NLP), CoNLL Multilingual Shallow Discourse Parsing shared task, SemEval (task on Semantic Evaluation)to name a few. As per (Luo et al., 2015), various low level tasks in NLP such as Parts-of-Speech Tagging (POST), chunking, parsing, Named Entity Recognition, are fundamental building blocks of complex NLP tasks such as Knowledge Base construction, text summarization, Question-Answering systems, and so on. Hence, the effectiveness of these low level tasks highly determine the performance of high end tasks. Error in low level tasks get propagated to high level tasks and degrading their overall performance. In this section, we will discuss about various sub-tasks in the field of Information Extraction.

\begin{center}
\includegraphics [scale=0.5]{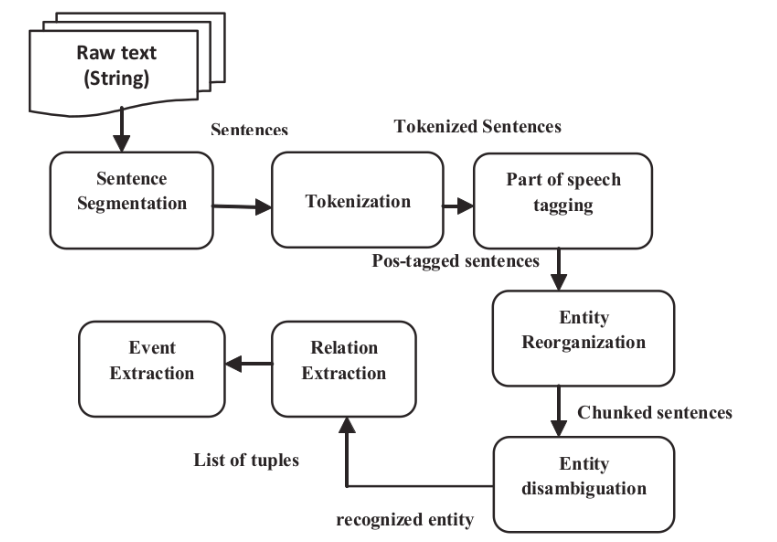}\\
\begin{small}Fig.1 General Information Extraction Architecture. Adapted from (Costantino et al., 1997)\end{small}
\end{center}

The effectiveness of various Information Extraction tasks down the pipeline highly depends upon pre-processing stages such as Tokenizer, Stemmer, Part-Of-Speech tagger,  and parser. Tokenizer extracts tokens from the text. Tokenizer can be treated as a classifier which classify tokens into 24 orthographic classes. Stemming process find the root (stem) of the word e.g. stem(reading)=read. Part-of-Speech tagger (POS-T) assigns tag to each word from various POS classes. e.g. Sam is a <proper-noun> and they is <personal-pronoun>. Noun Phrase Recognizer finds the noun phrases from the text. For e.g. "the president of USA", the president is noun phrase and it refers to a person, whereas USA represents noun phrase and refers to name of the country. Named Entity Recognizer, finally assigns particular named entity class from various classes such as: person, organization, location, date, time, money, percent, e-mail address and web-address. As Part-of-Speech tagging and Syntactic Parsing forms the building block and initial phase in the pipeline of various Information Extraction tasks, it is important to look at their role, state-of-the-art systems and how they effect downstream IE tasks.

\subsection{Parts-of-Speech (POS) tagging}
Part-of-Speech tagging labels unannotated words in natural language with Parts-of-Speech labels such as noun, verb, adjective, preposition,etc. (Chengyao Lv et al., 2016). Part-of-Speech tagging is fundamental step in various NLP tasks such as speech recognition, speech synthesis, machine translation, information retrieval, information extraction, and so on. Two factors that determine the tag of word are its lexical probability and its contextual probability (Voutilainen, 2003; Sun et al., 2008). Part-of-Speech tagging approaches can generally fall into two categories: Rule based approaches and statistical approaches. Rule based approaches apply language rules to improve the accuracy of tagging. The limitation of this approach lies in requirement of large annotated data which require expert linguistic knowledge, labor and cost. In order to overcome the shortcoming of this approach, (Brill,1995) proposed "transformation based approach" in which rules are automatically learned from corpora. On the other hand, statistical methods use Decision Trees (Magerman, 1995), Hidden Markov Model (Rabiner, 1989), Maximum Entropy classifier (Ratnaparkhi, 1996), Support Vector Machine (Gimenez and Marquez, 2004). Recently, deep learning based Part-of-Speech tagging has gained momentum.

\subsection{Parsing}
Numerous parser have been proposed by the Computational Linguistics community. These parsers can generally be divided into two broad categories based on their underlying grammatical formalism: Constituency parsers and dependency parsers. 

Constituency parsers (also known as tree bank parsers) produce syntactic analysis in the form of a tree that shows the phrases comprising the sentence and the hierarchy in which these phrases are associated. Constituency parsers have been used for pronoun resolution, labeling phrases with semantic roles and assignment of functional category tags.Constituency parsers overlook functional tags when training. Therefore, they cannot use them when labeling unseen text. 

Dependency parsers analyze the sentence as a set of pairwise word-to-word dependencies. Each dependency has a type that reflects its grammatical function. Dependency parsers model language as a set of relationships between words and construct a graph for each sentence, and each arc in the graph represents a grammatical dependency connecting the words of the sentence to each other.
\subsection{Named Entity Recognition (NER)}
In the Named Entity Recognition (NER) task, systems are required to recognize the Named Entities occurring in the text. More specifically, the task is to find Person (PER), Organization (ORG), Location (LOC) and Geo-Political Entities (GPE). For instance, in the statement "Michael Jordan lives in United States", NER system extracts Michael Jordan which refers to name of the person and United States which refers to name of the country. NER serves as the basis for various crucial areas in Information Management, such as Semantic Annotation, Question Answering, Ontology Population and Opinion Mining. 

\subsection{Named Entity Linking (NEL)}
Named Entity Linking (NEL) also known as Named Entity Disambiguation (NED) or Named Entity Normalization (NEN) is the task of identifying the entity that corresponds to particular occurrence of a noun in a text document. References to entities in natural text is quite ambiguous, because particular entity can refer to many mentions and on the other hand, particular mention can refer to many entities. For example, "Paris" can refer to entity mention location (GPE) i.e. city (Capital of France), or it can refer to entity mention person (PER) i.e. Paris Hinton (an actress). Hence, Entity Linking is the task of resolving named entity mentions to entries in structured Knowledge Base (KB) (Hackey et al., 2012). Named Entity Linking is quite important when information extracted from one document is to be integrated with information of same entity coming form various other sources. 

\subsection{Coreference Resolution (CR)}
Coreference Resolution is the task which determines which noun phrases (including pronouns, proper names and common names) refer to the same entities in documents (Kong et al., 2010). For instance, in the sentence, "I have seen the annual report. It shows that we have gained 15\% profit in this financial year". Here, "I" refers to name of the person, "It" refers to annual report and "we" refers to the name of the company in which that person works. Hence, coreference resolution plays vital role in tasks as natural language understanding, text summarization, information extraction, textual entailment, and so on.

\subsection{Temporal Information Extraction (Event Extraction)}
Temporal information extraction or event extraction refers to the task of identifying events (i.e information which can be ordered in a temporal order)in free text and deriving detailed and structured information about them, ideally identifying who did what to whom, where, when and why. Hence, temporal expression (also called timex)refers to the task of detecting phrases in natural language text that denote a unit of temporal entity in the form of an interval, a particular instance of time or certain frequency related to particular event. For instance, in the statement, "President Barack Obama yesterday addressed the issue of nuclear deals at White House". Here yesterday is a noun phrase which refers to temporal information. Temporal information is important where we want to extract structured information from natural language text according to some temporal criteria such as news, organization of events date-wise or biographies.

\subsection{Relation Extraction (RE)}
Relation Extraction is the task of detecting and classifying pre-defined relationships between entities identified in the text. In other words, it is way of transforming unstructured (free) text into structural form which can be used in web-search, question answering and lot more (Gardner et al., 2015). Given a sentence in free text \texttt{"Jodie Foster is an American acress who won Academy Award for movie The Silence of the Lambs"}, we can extract a fact such as \texttt{won(Jodie Foster, Academy Award)}. Pre-defined relations can be in the form of employee-of, Born-In, and Spouse-of where employee-of relation holds between particular person and an organization, Born-In relation holds between a person and a particular place where that person is born, and Spouse-of relation holds between two persons.

\subsection{Knowledge Base Reasoning and Completion}
There has been growing trend of constructing large Knowledge Bases(KBs) such as Freebase(Bollacker et al., 2008), DBpedia (Auer et al., 2007), YAGO (Suchanek et al., 2007), YAGO2 (Hoffart et al., 2011), Google Knowledge Graph (Dong et al., 2014), NELL (Mitchell et al., 2015). These KBs are currently in use for varied applications such as web-search, question-answering, decision support systems, digital assistants like Siri, Cortanan and Google Now. Though these KBs are large in size (having million of facts, billion of entities and million of relations), but they are error-prone and have lot many facts that are still missing. For instance, 71$\%$ of the persons described in Freebase have no known place of birth and 75$\%$ of them have no known nationality (Dong et al., 2014). Link prediction in Knowledge Bases (KBs) or Knowledge Graphs (KGs) aims at discovering new facts.

The goal of link prediction (also known as Knowledge Base Completion) is to determine the relationship between entities. There are various applications based on link prediction such as recommendation systems, Knowledge base completion and finding links between users in social networks. In recommendation systems, goal is to predict the rating of the movies which are not already rated and recommending it to users to have better user experience. Similarly, in Knowledge Base Systems, the goal is to check whether a particular triple not in the KB is likely to be true or not (Socher et al., 2011; Taskar et al, 2004). For instance, (Jodie Foster, LivesIn, ?) is a link prediction task where head (h) and relation (r) is given and we need to find the missing tail (t) from the triples already present in the KB. Hence, it make KB more robust to use for various applications such as question-answering, web/mobile search, social media analysis, recommendation systems, co-reference resolution, information retrieval and semantic parsing. 
\section{State-of-the-art-methods in Information Extraction}
The most important IE systems competitions are Message Understanding Conferences (MUC) starting with MUC-6 competition (Morgan et al., 1995).Current IE research seems to be heading in several directions, including partial parsing and automated knowledge acquisition (Riloff Ellen, 1997). The various approaches used in IE are broadly categorized into three main categories:
\subsection{Pattern matching based approach}
In this approach, extraction patterns are defined using a formalism called Regular Expressions (RE). These patterns can be easily matched directly with the given input text and the matched text is extracted, which corresponds to an occurrence of that entity. For example, if we want to extract corporate news, then we define simple Regular expressions with cue words such as Inc., Co., Company, Limited, Bank, Pty limited, Pvt. limited, and so on. 

Though it provides quick and easy process, but this approach has limitations as it is not possible to provide all the cue words related to particular domain. Moreover, company names such as ALDI, Woolworths, do not correlate with cue words. In order to make it more exploratory, Regular Expressions patterns are enriched by incorporating lexical information and incorporating special cases and domain knowledge. Despite its apparent limitations, this approach is widely used in practice.

\subsection{Gazetteer based approach}
Another approach is to make use of a pre-defined list of all possible values of an named entity, called a gazette or gazetteer. Gazette is only possible for only those named entities which have finite number of possible values. Though this approach is fast and accurate, but the limitations lies in preparing complete and accurate gazette.

\subsection{Machine Learning based approach}
In this approach, Machine Learning algorithms automatically learn the IE patterns by generalizing from a given set of examples. First we have to create a training data-set, which is a collection of documents in which all occurrences of named entities of interest are manually marked or tagged. ML algorithms such as Decision Trees, Naive Bayes classifier, Support Vector Machine (SVM), Conditional Random Fields (CRFs), Maximum Entropy (MaxEnt)use features such as word surrounding an occurrence of named entity. These supervised ML approach have limitations in terms of time and efforts required to create sufficiently large labeled training data-set. 

\subsection{Information Extraction Tools}
There are various publicly and commercially available tools for Information Extraction. 
\begin{itemize}
\item \textbf{Public IE tools:}GATE(General Architecture for Text Engineering), JULIE, OpenNLP (Apache OpenNLP -Java machine learning toolkit for NLP), Stanford NER, GExp, Mallet (Machine learning for language toolkit), Natural Language Toolkit (Suite of Python libraries for NLP), DBpedia Spotlight (Open source tool for Named Entity Recognition and Named Entity Linking) and OpenCalais (Automated IE web service from Thomson Reuters).
\item \textbf{Commercial IE tools:}Altensity, Open Calais, ClaraBridge, SAS Text Analytics, Business Objects, IBM Intelligent Minerand, Lingpipe.
\item \textbf{Specialized IE tools:}Ariadne Genomics Medscan Reader for biomedical documents, RINX for resumes.
\end{itemize}

\section{State-of-the-art-methods in various IE tasks}
\subsection{State-of-the-art in Named Entity Recognition (NER)}
Named Entity Recognition can be seen as a word-level tagging problem where each word in a sentence is mapped to a named entity tag. Features in the form of affix, capitalization, punctuation, output of syntactic analyzers (i.e. POS taggers, chunkers) and external resources in the form of gazetteers, word embeddings, and word cluster ids (Turian et al., 2010; Ratinov and Roth, 2009) are fed to classifier and output is the labeled tag in the form of person (PER), organization (ORG), location (GPE), etc. 

Eary NER systems were based on hand-crafted features. These hand-crafted features are fed to supervised Machine Learning (ML) systems in the form of rule-based systems or sequence labeling algorithms. Supervised ML approaches in NER task include: Hidden Markov Models (HMM), Support Vector Machines (SVM), Maximum Entropy models (MaxEnt), Decision Trees (DT), and Conditional Random Fields (CRFs). Among these, CRF provides better results as it takes context of near-by words into account. Then, research shifted towards using orthographic and language specific resources such as gazetteers. As these resources are costly to develop, requires expertise knowledge of linguistics, not transferable to new languages and domains, which limits their use. 

Recently, focus has been shifted on the use of semi-supervised learning (weakly supervised) for NER task. One such technique is "bootstrapping" which involves initial small degree of supervision in the form of providing set of seeds for starting the learning process (Burges et al., 1998).Siencnik et al., 2015 proposed NER using Word2Vec features. Qu et al., 2016 proposed transfer learning based NER in domains having quite less annotated data such as tweet messages. Kuru et al., 2016 proposed CharNER i.e. Character based named entity recognition. In this approach, instead of considering entire word as basic input feature, they took characters as the primary representation as in (Klein et al., 2003; Gillick et al., 2016). Secondly, they use a stacked bidirectional LSTM (Hochreiter and Schmidhuber, 1997)for final labeling task.

Tsai et al., 2016, proposed a cross-lingual NER model that is trained on annotated documents of one or more languages and applied to all languages in Wikipedia. Lample et al., 2016 proposed two neural architectures for NER based on bidirectional Long Short Term Memory (bidirectional LSTMs) and Conditional Random Fields (CRFs)and other model based constructs and labels segments using transition based approach inspired by shift-reduce parsers. The experimental results shows that LSTM-CRF model outperforms all state-of-the-art systems giving 90.94 F1-score on English NER task. 

\subsection{State-of-the-art in Named Entity Linking (NEL)}
The Named Entity Linking (NEL) also known as Named Entity Disambiguation or Named Entity Normalization task aims at automatically linking each named entity mention in a source text document to its unique referent in a target Knowledge Base (KB). For instance, from the statement: "Where would Pitt be without Jolie?", a named entity linker should link the entity mentions "Pitt" and "Jolie" to Brad Pitt and Angelina Jolie respectively which serves as a unique identifier for real people.

Two evaluations in particular has driven comparative work in the task of NEL: the TAC KBP (Knowledge Base Population)shared tasks and YAGO2 annotation of CoNLL-2003 NER data.

Early work on entity linking focused on treating as a separate task which was performed after the NER step. Typically, first NER system is run to extract entity mentions and then run an entity linking model to link mentions to a KB. As these two models are optimized separately, they are tractable and do not leverage information provided by one model to another. Hence, Luo et al., 2015 proposed JERL-Joint Entity Recognition and Linking, to jointly model Named Entity Recognition and linking tasks and capture the mutual dependency between them.
The SemEval-2015 multilingual Word Sense Disambiguation (WSD) and Linking task (Moro and Navigli, 2015) aims to promote joint research in the WSD and Entity Linking.

Recently, NEL systems uses both local and global information (Cucerzan, 2007; Ratinov et al., 2011; Alhelbawy and Gauzauskas, 2014). The local information measures the similarity between the text mention and an entity candidate in Knowledge Base, and global information measures how well the candidate entities in a document are connected to each other; with assumption that entities appearing in the same document should be coherent. Pershina et al., 2015 proposed Personalized Page Rank based random walk algorithm to disambiguate named entities. Avirup and Radu, 2016 proposed language independent entity linking (LIEL) system which trained on one language works for number of different languages. 

Although, Named Entity Recognition (NER), Coreference Resolution (CR), Cross-Document Coreference Resolution (CCR), and Named Entity Linking (NEL) involve close relations but they were not explored jointly. Chen and Roth, 2013; Zheng et al., 2013 shows that tight integration among these single sub-tasks provides promising results. Recently, several joint models have been proposed for CR-NER (Haghighi and Klein, 2010; Singh et al., 2013), CR-NEL (Hajishirzi et al., 2013), NER-CR-NEL (Durrett and Klein, 2014) and CCR-NEL (Dutta et al., 2015).

\subsection{State-of-the-art in Coreference Resolution (CR)}
Coreference Resolution (CR) is the process of linking together multiple expressions of a given entity. Coreference Resolution plays significant role in making Information Extraction, text summarization and Question-Answering systems more robust. Coreference Resolution plays vital role in Relation Extraction (RE) task which aims at finding pairs (tuples) of entities that satisfy some defined relation. For instance, "Sam is a man of great talent. He works for Macquarie University". Here, if we ask question: "Where does Sam work?", we can't find a direct statement that answers it. But if we establish relation <work> between He and Macquarie University, and if we find that Sam and He are in coreference, we can conclude that works(Sam, Macquarie University) and we get the answer to the query. 

Early work in coreference resolution started with levaraginng lexical and syntactic attributes of the noun phrases such as string matching, the distance between coreferent expressions, name alias (Soon et al., 2001; Ng and Cardie, 2002). Shane Bergsma et al., 2006 presented approach based on syntactic paths for pronoun resolution. Young et al., 2006 proposed a kernel based method to mine syntactic information from parse trees. Zhou et al., 2008 proposed context-sensitive convolution tree kernel for pronoun resolution. Heuristic rule-based approaches which concentrate on designing heuristic rules to identify the anaphoric relation between noun phrases were also used for coreference resolution (Lappin et al., 1994; Kennedy et al., 1996; and vieira et al., 2000).

After this, machine learning approaches became more prevalent which considers it as binary classification task (Cardine et al., 2008). At first, ML approaches focused on exploiting lexical, grammatical and syntactic features, then researchers shifted focus on using semantic information. Huang et al., 2010, proposed coreference resolution in biomedical full-text articles with domain dependent features. They showed that using appropriate domain dependent features can improve coreference resolution results. Results showed 76.2\% Precision, 66\% Recall and 70.7\% F1-measure which is 13.8\% improvement against not using domain dependent features. Among the ML approaches, one aspect of coreference resolution is to directly utilize the parse trees as a structured feature and use a kernel based method to automatically mine the knowledge embedded in the parse trees. Kong et al., 2010 proposed a dependency driven scheme to dynamically determine the syntactic parse tree structure for tree kernel based anaphoricity determination in coreference resolution. Experiments on ACE-2003 Corpus demonstrates that this approach yield improved F1-score by 2.4, 3.1 and 4.1 on NWIRE, NPAPER and BNEWS domains.

Currently, several supervised entity coreference resolution systems can be categorized into three classes: mention pair models (McCarthy et al., 1995), entity-mention models (Yang et al., 2008; Haghighi and Klein, 2010; Lee et al., 2011) and ranking models (Yang et al., 2008; Durrett and Klein, 2013; Fernandes et al., 2014). Among all the three main approaches, ranking models provides state-of-the-art performance for coreference resolution. Recently, trend is shifting towards using unsupervised learning algorithms for coference resolution. Haghighi and Kelin, 2007 presented a mention pair non-parameteric fully-generative Bayesian model for unsupervised coreference resolution. Ng et al., 2008 proposed probabilistic induced coreference partitions via Expectation Maximization (EM) clustering. Poon and Domingos, 2008 proposed an entity-mention model to perform joint inference across mentions using Markov Logic. Ma et al., 2016 proposed an unsupervised generative ranking model. Until 2015, all state-of-the-art coreference systems operate solely by linking pairs of mentions together (Durrett and Klein, 2013; Martschat and Strube, 2015; Wiseman et al., 2015). Recently, Kevin Clark and Chris D.Manning, 2016 use agglomerative clustering approach which treats each mention as a singleton cluster at the outset and then repeatedly merging clusters of mentions deemed to be referring to the same entity. Such system takes advantage of entity level information i.e. features between cluster of mentions instead of between just two mentions. Moreover, Cross-document coreference resolution (Gao et al., 2010) is at the research forefront in research community. Event Coreference Resolution can be applied within a single document or across multiple documents and is crucial for various NLP tasks including topic detection and tracking, IE, Question-Answering, and textual entailment (Bejan and Harabagiu, 2010). In comparison to entity coreference resolution i.e. identifying and grouping noun phrases that refers to the same discourse entity, event coreference resolution is more complex as single event can be described using multiple event mentions. More recently, SemEval-2010 Task 1 was dedicated to coreference resolution in multiple languages. In CoNLL-2011 shared task, participants had to model unrestricted coreference in the English language OntoNotes corpora and CoNLL-2012 shared task involved predicting coreference in three languages-English, Chinese and Arabic. Recent work is dominated by machine learning approaches. There are many open-source platforms based on machine learning for coreference resolution such as BART (Verseley et al., 2008), the Illinois Coreference Package (Bengston et al., 2008) and the Stanford CoreNLP (Manning et al., 2014).

\subsection{State-of-the-art in Temporal Expression Extraction}
Often we find that information in newspaper stories, stories, biographies, corporate news and documentaries is temporally ordered. Temporal information extraction (also known as event extraction) finds applications in temporal Question-Answering systems, Machine Translation and document summarization. Temporal information extraction involves the identification of event-time, event-document and event-event relations from free text (Kolya et al., 2010). Most of the event extraction systems consists of three sub-tasks (Amami et al., 2012):
(1). Pre-Processing: It provides syntactic and semantic analysis of text as an input to event detector module. Steps include sentence splitting, tokenizing, Part-of-Speech tagging, and parsing.
(2). Trigger detection: It assigns each token to an event class. Here the task is to identify individual words in the sentence that acts as an event trigger word and assigning the correct event class to each of the determined trigger. (3). Argument detection: This step consists of finding all participants in an event and assigning the functional role to each of the determined participants in an event. 

Shared tasks in the form of TempEval which aims at finding temporal relations (Verhagen et al., 2007) promote research in event extraction. Several methods have been proposed to extract events from text. Early approaches include Pattern-matching which try to use context information for finding relations between entities. Boguraev et al., 2005; Mani et al., 2006; and Chambers et al., 2007 proposed machine learning based approaches for temporal relation identification. Kolya et al., 2010 used CRF based approach for event-event relation identification. They used some of the gold standard TimeBank features for events and times for training the CRF. Various features used were: Event class, Event stem, Event and time string, POS of event terms (Adjective, Noun, Verb, Adverb, Prep), event tense (Present, Past, Future, Indefinite, None), event aspect (e.g. progressive, perfective, prefective-progressive, or  none), event polarity (positive,negative, neutral), event modality, type of temporal expression, temporal signal and temporal relation between document creation time and the temporal expression in the target sentence(greater than, less than, equal, or none). They reported Precision, Recall and F-score values of 55.1\%, 55.1\% and 55.1\% respectively under strict evaluation scheme and 56.9\%, 56.9\% and 56.9\% respectively under relaxed evaluation scheme. TimeBank Corpus (Pustejovsky et al., 2003) has largely promoted the development of temporal relation extraction. Recently, hybrid approaches based on hand coded rules along with supervised classification models using lexical relation, semantic and discourse features are proposed by D'Souza and Ng, 2013;and Chambers et al., 2014. Mirza et al., 2016 proposed sieve based system to perform temporal and causal relation extraction and classification.

\subsection{State-of-the-art in Relation Extraction}

Relation Extraction enables broad range of applications including Question-Answering, Knowledge Base Population, and so on. The relation extraction task can be divided into two steps: detecting if a relation utterance corresponding to some entity mention pair of interest occurs and classifying the detected relation mentions into some predefined classes. There are two types of Relation Extraction systems: Closed domain relation extraction systems consider only a closed set of relationships between two arguments. On the other hand, Open-domain relation extraction systems use an arbitrary phrase to specify a relationship. In order to extract relations, the sentences in the free text are analyzed using Parts-of-Speech tagger a dependency parser and a Named Entity Recognizer. Early RE approaches can be broadly divided into two categories: feature based methods (Kambhatla, 2004; Boscher et al., 2005; Zhou et al., 2005; Grisham et al., 2005; Jiang and Zhia, 2007; Chan and Roth, 2010; Sun et al., 2011; Nguyen and Grisham,2014) and kernel-based methods (Zelensko et al., 2003; Culotta and Sorensen, 2004; Bunescu and Mooney, 2005; Sun and Han, 2014). Lexico-Syntactic Patterns (LSP) or Hearst Patterns at word level, phrase level and sentence level as syntactic features for relation extraction. Alternate approach is to use semantic features (Alicia,2007; Hendrickx et al., 2007)of a pair of words extracted from lexical resources like WordNet(Fellbaum, 1998). Hybrid approaches combines syntactic patterns with semantic features of the constituent words(Claudio, 2007; Girju et al., 2005). Tesfaye et al., 2016 proposed such hybrid approach getting semantic information extracted from the Wikipedia hyperlink hierarchy of the constituent words. Relation extraction systems (Feldman et al., 2008; Etzioni et al., 2008) utilize pipeline of processes such as Part-of-Speech tagging, phrase detection, Named Entity Recognition and Coreference Resolution. Recently, (Boden,et al., 2011)shows that higher precision on RE task is ensured with deep dependency parsing or semantic role labeling/analysis.

Many approaches to relation extraction use supervised machine learning techniques (Soderland et al., 1995; Califf and Mooney, 1997; Lafferty et al., 2001), but these methods require large human annotated training corpus. In order to overcome the problem of labeled data-set, researchers developed Distant Supervision (DS) in which a knowledge base such as Wikipedia or Freebase is used to automatically tag training examples from the text corpora(Mintz et al., 2009).Specifically, distant supervision uses the KB to find pair of entities $E_1$ and $E_2$ for which a relation $R$ holds. Distant Supervision then makes assumption that any sentence that contains a mention of both $E_1$ and $E_2$ is a positive training instance for $R(E_1,E_2)$. Unfortunately, distant supervision technique leads to large proportion of false positive training instances. To address this shortcoming, there have been attempts to model the relation dependencies as Multi-Instance Multi-Class (Bunescu and Mooney, 2007; Riedel et al., 2010) leading to state-of-the-art relation extraction results. Based on this approach, Hoffmann et al., 2011 proposed MultiR system for relation extraction and Surdeanu et al., 2012 proposed MIML-RE (Multi-Instance Multi-Labeling Relation Extraction) system. Additionaly, other relation extraction techniques include Universal Schemas (Riedel et al., 2013; Deep learning (Nguyen and Grisham, 2014). 

Recently, researchers have explored the idea of augmenting distant supervision with a small amount of crowdsourced annotated data in order to improve relation extraction performance (Angeli et al., 2014; Zhang et al., 2012; Pershina et al., 2014). With the growing popularity of Word2Vec model , Hashimoto et al., 2015 proposed embedding based model for relation extraction. Recent trend in relation extraction is to use deep learning techniques such as Convolutional Neural Networks (CNN) (dos Santos et al., 2015), Recurrent Neural Networks (RNN) (Socher et al., 2012), tree structured LSTM-RNN (Tai et al., 2015),and  LSTM (Miwa et al., 2016). Lin et al., 2016 proposed attention based model in distant supervision relation extraction which proves to provide better state-of-the-art results in relation extraction.There has been growing interest in multi-lingual relation extraction. Gamallo et al., 2012 presented a dependency parser based on Open Relation Extraction system for Spanish, Portugese and Galician. Recently, Tseng et al., 2014 proposed an Open RE (Open Relation Extraction) for Chinese that employs word segmentation, POS tagging and dependency parsing. Lewis and Steedman, 2013 learn clusters of semantically equivalent relations across French and English by creating a semantic signature of relationship by entity-typing. These relations are extracted using CCG (Combinatory Categorical Grammar)parsing in English and dependency parsing in French. Gerber and Ngomo, 2012 described a monolingual pattern extraction system for RDF predicates that uses pre-existing KB for different languages.

Past two decades have witnessed a significant advancement in extracting binary domain-dependent relations, but modern question-answering and summarization systems have triggered interest in capturing detailed information in a structured and semantically coherent fashion, thus motivating the need for complex n-ary relation extraction systems (Khirbat et al., 2016). McDonald et al., 2005 proposed n-ary relation extraction system that factorize complex n-ary relation into binary relations, representing them in a graph, and tried to reconstruct the complex relation by making tuples from selected maximal cliques in the graph. Then, each relation is classified using Maximum Entropy (MaxEnt) classifier. Recently, Li et al., 2015 make use of lexical semantics to train a model based on distant supervision for n-ary relation extraction. Khirbat et al., 2016 proposed algorithm for extracting n-ary relations from biographical data which extracts entities using Conditional Random Fields (CRF) and n-ary relations using Support Vector Machine (SVM)from two manually annotated data-sets which contains biography summaries of Australian researchers.

Recently, relation extraction models based on deep learning have achieved better performance than conventional relation extraction models that rely on hand-crafted features. Li et al., 2016 proposed a pre-training method that generalizes well known Seq2Seq model for deep relation extraction models in order to reduce the need of annotated data. Shen et al., 2016 proposed attention based Convolutional Neural Network (CNN) architecture which uses word embedding, part-of-speech tag embedding and position embedding information. They showed that word level attention mechanism is able to better determine parts of the sentence which are most influential w.r.t two entities of interest.

\subsection{State-of-the-art in Knowledge Base Reasoning and Completion}
Most of the real-world data we see around us is inherently relational. Whether it is social networks, gene-protein interactions, hyper linking pages for world wide web, clustering of documents based on particular topic, or author-publication citations. Relational machine learning (SRL) refer to set of methods which are used for statistical analysis of relational or graph-structured data (Nickel et al., 2015).  SRL aims at processing data in the form of sets, graphs or similarity complex structures, where the standard representation are not used (Blockeel et al., 2006). SRL is well applied in many specialized areas such as web-mining, social network analysis, temporal and spatial analysis which are highly inherently relational in nature. Prior work on multi-relational learning can be categorized  into  three  categories:  (1)  statistical  relational  learning  (SRL),  such  as  Markov-logic networks, which directly encode multi-relational graphs using probabilistic models; (2) path ranking methods, which explicitly explore the large relational feature space of relations with random walk; and (3) embedding-based models, which embed multi-relational knowledge into low-dimensional representations of entities and relations via tensor/matrix factorization, Bayesian clustering framework, and neural networks.  SRL models try to predict unseen or future relations between entities based on the data already present in the Knowledge Bases (KBs).

\subsubsection{Probabilistic Latent Variable models}
Probabilistic latent variable models conditions the probability distribution of the relations between two entities on the latent attributes entities and all relations are considered conditionally independent given the latent attributes (Minervini et al., 2016). These models are quite similar to Hidden Markov Models (Xu et al., 2006; Koller and Friedman,2009) which allows the information to propagate through the network of interconnected latent variables. Wang and Wong (1987) proposed Stochastic Block Model(SB) which associates a latent class variable with each entity. The SB model is further improved using Bayesian non-parametrics in order to find the optimal number of latent class as given in Infinite Relational model (Kemp et al., 2006; Xu et al., 2006). These models are further extended in Infinite Hidden Semantic Model (Rettinger et al., 2009), Mixed Membership Stochastic Block Model (Airoldi et al., 2008) and Non-parametric Latent Feature Relational Model (Miller et al., 2009) which was based on Bayesian non-parametrics.

Though probabilistic latent variable models gave promising link prediction results, but showed limitations on scaling on large Knowledge Graphs because of the complexity of the probabilistic inference and learning (Koller and Friedman, 2009). As a consequence, these models do not result to be fully appropriate for modeling large KBs.

\subsubsection{Embedding models}
Embedding models learns vector space representation for the entities and relations in the KB and use these representations to predict the missing facts. Latent feature models do reasoning over the Knowledge Bases (KBs) via latent features of entities and relations. The intuition behind such models is that the relationship between two entities can be derived from the interactions of their latent features.Early work focused on Matrix Factorization approaches. Recently, trend is moving towards using tensors for KB completion task. Tensors are multidimensional arrays which represents multi-relational data quite easily. Tensor decomposition have been widely used in the fields of Chemo-metrics and psycho- metrics but now they are receiving huge attention by machine learning community as tensors have capability to represent social analytics data, genomic data, seismic data, brain signals and other multi-dimensional data. Similar to matrix factorization, tensor decomposition approaches use component matrices to represent the latent space of the corresponding dimension and a core tensor to represent the interactions among these component matrices. Predictions are made via generalized matrix operations to reconstruct the tensor from the latent factors. Though tensors provide greater descriptive flexibility of multi-relational data but this flexibility comes at the cost of huge computation which is quite low in matrices. Tensors are sometimes quite similar to extending Singular Value Decomposition (SVD) approach from 2-D matrices to higher dimensions. Nickel et al., (2013) represents knowledge base having $n$ entities and $m$ binary relations in the form of a three-way tensor $K$ of size $n \times n \times m$. This method approximates each slice of the KB tensor $K$ as a product of an entity matrix, a relation matrix, and the entity matrix transposed. Various other tensor decomposition methods such as CANDECOMP/PARAFAC (CP) Decomposition and Tucker Decomposition (see Kolda and Bader, 2009) are also employed for various information extraction tasks. Following RESCAL model, lot many extensions were proposed which takes weighted tensor decomposition (London et al., 2013), coupled matrix and tensor factorization (Papalexakis et al., 2014) as well as collective matrix factorization approaches (Singh et al., 2008). In order to further extend the functionality, leveraging relational domain knowledge along with tensor decomposition (Chang et al., 2014) has been proposed.  \\ 

One of the promising breakthroughs for link prediction in KBs is embedding a knowledge graph into a continuous vector space while preserving certain information of the graph (Socher et al., 2013, Bordes et al., 2013a, Weston et al., 2013, Chang et al., 2013). All of these models differ in their projection of embeddings in vector space, score function f(h,r,t) and algorithms to optimize their margin based objective function.  One of the most important model which leads to hundreds of paper in the field of KBC is TransE (Bordes et al., 2013). The basic idea of TransE is to represent relationship between two entities corresponds to translation between the embeddings of the entities, i.e h + r ≈ t when the triple (h,r,t) holds true. This indicates that the tail entity (t) should be the nearest neighbor of (h, r). Hence, TransE assumes the score function high if (h,r,t) holds and low if (h,r,t) do not holds true. In order to address issues for N-to-1, 1-to-N and N-to-N relations in TransE, various extensions TransH model (Wang et al., 2014), STranE (Nguyen et al., 2016), TransR/CTransR(Lin et al., 2015)were proposed. For each relation r, TransH models the relation as a vector on a hyperplane with $W_r$ as a normal vector. Thus, entity embeddings h and t are first projected to the hyperplane of $W_r$, denoted as $h\bot$ and $t\bot$.  Also, STransE model integrates SE model with TransE model to have two relation specific matrices $W_{r,1}$ and $W_{r,2}$ to identify relation-dependent aspects of both h and t. TransR/CTransR model projects entities and relations in different embedding space in order to capture various contexts of the same relation with different entities.  PTranE(Lin et al., 2015a)takes multiple step relation paths between entities indicating their semantic relationships. TransD(Ji et al.,2015) considers the multiple types of entities and relations simultaneously, and replaces transfer matrix by the product of two projection vectors of an entity relation pair. TransG(Xaio et al., 2015)addresses the issue of multiple relation semantics that a relation may have multiple meanings revealed by the entity pairs associated with the corresponding triples. KG2E (He et al., 2015) uses Gaussian embedding to model the data uncertainty. Though it has good results on 1-to-N and N-to-1 relations but it do not provide reasonable good results on 1-to-1 and N-to-N relations. Recently, (Minervini et al., 2016) found that adaptive learning rate plays significant role for learning model parameters. They tried various optimization algorithms such as SGD, AdaGrad (Duchi et al., 2011), AdaDelta (Zeiler, 2012, Adam (Kingma et al., 2015) and found that AdaGrad with its adaptive learning rate provides the best link prediction results.\\

Completeness, accuracy and data quality are important parameters that determine the usefulness of knowledge bases and are influenced by the way knowledge bases are constructed (Nickel et al., 2015). Recent work based on modeling relation paths between entities by (Neelakantan et al., 2015; Gardner and Mitchell, 2015; Luo et al., 2015; Lin et al., 2015a; Guu et al., 2015; Garcia-Duran et al., 2015; and Toutanova et al., 2016) showed richer information and improved relation prediction. TransE-NMM (Nguyen et al., 2016) takes account of neighborhood entities and relation information of both head and tail entities in each triple. Experiment results on TransE-NMM model gives better results on triple classification and link prediction. Moreover, RTransE (Garcia-Duran et al., 2015), PTransE (Lin et al., 2015a) and TransE-COMP(Guu et al., 2015) are all extensions of TransE model. 

Link prediction or Knowledge Base Completion and relation extraction are complementary to each other. Recent approaches in information extraction combine text-based extraction models with link prediction models to infer new relational facts to achieve significant improvement over the individual approaches. In (Weston et al., 2013), TransE link prediction model and text based relation extraction model are combined to rank candidate facts and promising improvement in relation extraction task is achieved. Similar improvement has been seen over TransH (Wang et al., 2014) as well as TransR(Lin et al., 2015). Link prediction techniques on KB can make prediction given a single textual predicate as an example, and thus can be used directly on the relation extraction task as shown by (Riedel et al., 2013) using matrix factorization for relation extraction. Apart from this, current research in Knowledge Bases aims at constructing a KB that is sufficiently accurate, has broad coverage, and gets evolved over time. There are three main challenges at the forefront of Knowledge Base reasoning and completion research. These are: (1). How to create a KB from scratch with good accuracy and coverage by leveraging pages on the web. (2). How to solve the performance challenges at the web-scale, and (3). How to merge knowledge extracted from different sources or using different methods, to achieve both higher coverage and better quality.

\section{Integrating Information Extraction (IE) with Information Retrieval (IR) Systems}
The task of Information Retrieval (IR) system is to provide a list of documents which are most relevant in response to user query. Search engines like Google or Bing, use web crawlers to gather documents and create a massive index of these documents by noting which words occur in which document, and answer queries by identifying documents that contains these keywords. Then it uses intelligent ranking algorithms (like PageRank) to put most likely ones at the top (i.e. in decreasing order). But still these search engines have limitations as they provide only documents and no specific answer to the query. Moreover, IR systems are still keyword (string matching)based. To overcome these limitations, NLP research community has moved towards integrating Information Extraction technology with Information Retrieval technology to make search engines more reliable, accurate, specific to user queries and having high semantic understanding. Though, user queries can be answered as a Question-Answering problem where we can provide direct answer to user queries, but in order to accomplish this goal, we need highly structured Knowledge Base and better reasoning/inference techniques (Paik et al., 1997). Thus it seems that that the "Semantic Web" paradigm presented by Tim Berners Lee (Lee et al., 2001), according to which the Internet should become a structured container for semantic, machine interpretable model of knowledge and information is still far to be achieved.

\section{Applications of IE}
Information Extraction acts as a key technology in various Natural Language Processing (NLP) applications such as Machine Translation, Question-Answering, Text Summarization, Opinion mining, etc. The research in the field of IE is in infancy and it is of great significance to information end-user industries such as finance, banks, publishers, governments, etc. 

\begin{itemize}
\item \textbf{Question-Answering System:}Recent advances in constructing large scale Knowledge Bases (KBs) have enabled new Question-Answering systems to return exact answer from a KB. State-of-the-art methods in Question-Answering Systems are divided broadly into three main categories: Semantic parsing, information retrieval and embedding based. Semantic parsing (Cai and Yates, 2013; Berant et al., 2013; Kwiatkowski et al., 2013; Berant et al., 2014; Fader et al., 2014) aim to learn semantic parsers which parse natural language questions into logical forms and then query KB to lookup answers. Though, this approach put constraint on training as it is difficult to train model at large scale because of their complexity of their inference, they tend to provide deep interpretation of the question. Information retrieval based systems retrieve a set of candidate answers and then conduct further analysis to rank them according to some specific criteria. Embedding based approaches (Bordes et al., 2014b; Bordes et al., 2014a) learn low dimensional vectors for words and knowledge base constituents, and use sum of these vectors to represent questions and their candidate answers. 

\item \textbf{User Profile Extraction:}With the rise of social media platforms such as Twitter, Facebook, Instagram and Tumblr. They provide fast source of information as compared to traditional sources such as news, articles, and magazines. Information Extraction technology is used to extract information from these social media in the form of personal information (profile, age, gender, address, political orientation, business affinity, user interests), social information (family members, friends, co-workers, communities involved) and travel information (current location, travel history, vacation, etc.)
\item \textbf{Auto E-mail reply and Auto-Calender:}We send are receive large number of emails on daily basis. Majority of these emails have to be replied in Yes/No mode and rest of the emails contains information about various upcoming events (like upcoming meetings, seminars, showcases, conferences, etc.). Reading, indexing, replying and putting all the information in the calender format and finally adding to calender is quite tedious and time consuming task. Information Extraction technology, specifically event extraction techniques automatically extract events from emails and add them to the calender.
\item \textbf{Resume Processing:}Extraction of information from resumes is quite important for job/placement agencies. Manually reading individual resume, checking their prospects as per job requirements and replying for interview call to every individual is really messy. Using Information Extraction technology, resume are processed and replied automatically. As the format of resume is not fixed and each job seeker makes it as per its interest and requirement of industry, it makes this task more difficult. Agencies use "resume filters" to filter out unwanted resumes and make their task easier. Simple keyword search (string matching) based techniques may result in ignoring prospective candidates and hence, high level information extraction technology is used to have more robust systems.
\item \textbf{Classified Advertisements:}The advertisements posted in newspapers, news bulletin, websites and shopping stores (both online and offline) have unstructured format in the form of natural text. Extracting attributes using Information Extraction technology to have structured information depending upon desired attributes is highly beneficial as compared to searching through natural text. Hence, IE technology is used to populate the rows in a relational database with values for certain attributes of interest as well as to classify various advertisements into some pre-defined classes like cars, shoes, kitchenware, electronic, apparel, and so on.

\item \textbf{Customer Services:} Companies receives lot of unstructured information in the form of emails, chat transcripts, support forum discussions, and daily transactions reports. So, companies use IE tools to retrieve relevant information from these sources, classifying them and forming structured database which makes the task much easier as compared to searching and retrieving manually.

\item \textbf{Business Intelligence:} Due to dynamic nature of businesses and to remain ahead from competitors, organizations acquire market information in the form of competitors pricing, public opinion, market study, promotions by competitors, and business strategies using web information extraction technology. Top Business Intelligence tools available in the market are: SAS Analytics Pro, IBM Cognos, Microsoft Power BI, Tableau, and SAP. These BI tools acts as a data management tools by collecting data from various sources, cleaning data, organizing data in proper format and structure and preparing databases for better visualization and reporting. Hence, better data management helps companies to have better decision making which is vital to remain in business.

\item \textbf{Building citation database:} With increasing number of scientific papers online, linking references automatically without manual intervention is quite necessary.  Citations helps the scientific community to properly credit the information they use. Building database of citations helps user to search, count number of citations and cross-referencing of scientific papers. Hence, IE technology is used to build various citation building  systems such as Citeseer, Google Scholar, DBLP, Scopus, Thomson Reuter, etc. 

\item \textbf{Social Analysis:} Social media platforms such as Facebook, Twitter, Instagram and blogs are acting as new style of sharing and exchanging information. Extracting information from social media poses various challenges as text is noisy, quite informal, and short in context. To make use of this vast information, we need better IE tools which meet the listed challenges. Social media analysis has gained lot of attention in IE research community as it provides up-to-date information compared to conventional sources such as news.
\end{itemize}

\section{Current challenges and future research}
\subsection{Open Information Extraction (OpenIE)}
OpenIE have been drawing more and more attention from research community to enhance and scale IE systems by utilizing large, complex and heterogenous data; and extracting all meaningful relations and events without any restrictions. Current challenges includes extending the capability of OpenIE to handle n-ary relations and even nested extractions, dealing with multiple languages, extraction of temporarily changing information and to distinguish between facts, opinions and misinformation on the web. Moreover, research is in the direction of integrating IE and summarization. IE extracts important information in the form of named entities, events, relations, and then this information is fed to summarization template which provides summary of the actual text.

\subsection{BioIE}
Though lot of progress has been made in NLP to handle common unstructured text, less attention has been given to bio-medical text. Bio-medical text in the form of patient discharge summaries, doctor's prescriptions, scientific publications provides various challenges to standard IE techniques. Hence, BioIE is important to various applications in healthcare industry including clinical decision support, integrative biology, bioinformatics, biocuration assistance, and pharmacovilance. State-of-the-art IE techniques still lack upto human level performance in health-care industry. Hence, future research is to explore and develop IE techniques for health-care industry.

\subsection{Business Analytics}
Business Intelligence and Analytics refers to set of tools and systems that analyses data to help an organization to better understand its business and markets,  and provides a way of better decision making. Business Analytics heavily relies on various advanced data collection, extraction, analysis and summarization technologies. Current state-of-the-art business analytics tools can work on clean and stored data. As, text generated by users is often grammatically wrong, error-prone, and abbreviated, future research is in the direction to have IE techniques that can work on noisy unstructured data and having capability to handle stream data in order to have processing in real time.

\subsection{Text IE in Images and Videos}
Images and videos contains lot of text data which acts as an useful information for automatic annotation, indexing and structuring of images and videos. Text data present in images and videos can acts as a cue stating what image or video is about. Hence, extracting text data from images and videos can greatly help in retrieval systems and indexing. Text information extraction from images and videos involve various steps such as detection, localization, tracking, extraction, enhancement and recognition of text. Each step poses various challenges as there lies lot of variations in text due to differences in size, style, orientation, contrast, and complex background. The research in this area is still in its nascent stage and scientists are working to apply better computer vision and natural language processing techniques to improve its performance.
 
\subsection{Web Harvesting}
As the amount of information on the web grows, it becomes harder to keep track of and to effectively make use of this information. How one can find the information one is looking for in a useful format, easily and quickly is the current challenge. Search engines scan the documents and return the URLs. They also find and return meta description and meta keywords embedded in web pages, but do not provide the exact information as per user requirements. The general process of finding information by a user is: (1). Scan the content until we find the information, (2). Mark the information by selecting or highlighting it, (3). Switching to another program such as spreadsheet, word processor or database, and (4). Pasting the information into that application. Web Harvesting tools automate the reading, copying and pasting necessary to collect information for analysis. Hence, future research is to develop web harvesting tools which acts as an end-to-end information extraction systems.

\section{Conclusion}
Information Extraction field has progressed with variety of information extraction techniques such as Open Information Extraction (OpenIE), semi-structured extraction via infoboxes and various KBs such as Google Knowledge Graph, Microsoft Satari, YAGO, DBPedia, NELL and Probase. 

There is considerable excitement in NLP community at the prospects of Information Extraction technology due to rise of social media platforms such as Twitter, Facebook, Instagram, and so on. Social media reacts to world events faster than traditional news sources, and its sub-communities play close attention to topics that other sources might ignore. But analyzing the text in social media is challenging as the text is short, language is informal, capitalization is inconsistent, spelling variations and abbreviations run rampant. Moreover, research is shifting towards joint models that learn two or more tasks. McCallum et al., 2003 proposed joint POS tagging and chunking, Finkel and Manning, 2009 proposed joint model for parsing and NER together, Yu et al., 2011 worked on jointly entity identification and relation extraction from Wikipedia, Sil, 2013 proposed joint NER and Entity Linking, Surdeanu et al., 2008 jointly optimized parsing and semantic role labeling, Wang et al., 2015 proposed joint model for Information Extraction and KB completion based on statistical relational learning (Getoor and Taskar, 2007).

Recently, deep learning has been widely used in various areas including Computer Vision, Speech Recognition, and Natural Language Processing. In NLP tasks, deep learning has been successfully applied to Part-of-Speech tagging (Collobert et al., 2011), Sentiment Analysis (dos Santos and Gatti, 2014), Parsing (Socher et al., 2013), Machine Translation (Sutskever et al., 2014). Moreover, attention based models based on deep learning has attracted lot of interest in research community. It has been successfully applied to Image Classification (Mnih et al., 2014), Speech Recognition (Chorowski et al., 2014), Image Captioning (Xu et al., 2015), Machine Translation (Bahdanau et al., 2014), and Part-of-Speech tagging (Barrett et al., 2016). Future research is in direction to have web harvesting tools for end-to-end information extraction as well as improving existing techniques to handle noisy data from social media.\\

\begin{Large}\textbf{References}\end{Large}
\vspace{0.1cm}

\textbf{[Costantino et al., 1997]} Marco Costantino, Richard G Morgan, Russell J. Collingham, Roberto Garigliano, Natural Language Processing and Information Extraction: qualitative analysis of financial news articles, 1997.
\vspace{0.1cm}

\textbf{[Morgan et al., 1995]} R.Morgan, R.Garigliano, P.Callaghan, S.Poria, M.Smith, A.Urbanowicz, R.Collingham, M.Costantino, C.Cooper, and the LOLITA group, University of Durham: Description of the LOLITA system as used in MUC-6, in Sixth Message Understanding Conference (MUC-6), Morgan Kaufman, November, 1995.
\vspace{0.1cm}

\textbf{[Neil et al., 1998]} Paul O' Neil and Woojin Paik, The Chronological Information Extraction System (CHESS), IEEE, 1998.
\vspace{0.1cm}

\textbf{[Ellen Riloff, 1997]} Riloff Ellen, Information Extraction as a stepping stone toward story understanding, in Understanding Language Understanding: Computational models of Reading, edited by Ram and Moorman, MIT Press, 1997.
\vspace{0.1cm}

\textbf{[Dzunic et al. 2006]} Zoran Dzunic, Svetislav Momcilovic, Branimir Todorovic and Miomir Stankovic, Coreference Resolution Using Decision Trees, 8th Seminar on Neural Network Applications in Electrical Engineering, NEUREL-2006.
\vspace{0.1cm}

\textbf{[Bikel et al., 1997]} M.D.Bikel, S.Miller, R.Schwartz and R.Weischedel, Nymble: A high performance learning name-finder, In Proceedings of the Fifth Conference on Applied Natural Language Processing, 1997, pp. 194-201.
\vspace{0.1cm}

\textbf{[Fang et al., 2008]} Kong Fang, Liyancui, Zhou Guodong, Zhu Qiaoming, Qian Peide, Using Semantic Roles for Coreference Resolution, International Conference on Advanced Language Processing and Web Information Technology, IEEE, 2008.
\vspace{0.1cm}

\textbf{[Soon et al., 2001]} W.M Soon, H.T. Ng and Lim, A Machine Learning approach to coreference resolution of noun phrase, Computational Linguistics, 2001.
\vspace{0.1cm}

\textbf{[Ng and Cardie, 2002]} V.Ng and C. Cardie, Improving machine learning approaches to coreference resolution, Proceedings of the 40th Annual Meeting of the Association  for Computational Linguistics, 2002.
\vspace{0.1cm}

\textbf{[Bergsma and Lin, 2006]} S.Bergsma and D.K.Lin, Bootstrapping path-based pronoun resolution, COLING-ACL, 2006.
\vspace{0.1cm}

\textbf{[Yang and Su, 2007]} X.F.Yang and J.Su, Coreference Resolution using Semantic Relatedness Information from Automatically Discovered Patterns, ACL, Prague, Czech Republic, June, 2007.
\vspace{0.1cm}

\textbf{[Zhou et al., 2008]} G.D.Zhou, F.Kong and Q.M.Zhu, Context-sensitive convolution tree kernel for pronoun resolution, IJCNLP, 2008.
\vspace{0.1cm}

\textbf{[Gao et al., 2010]} Sanyuan Gao, Si Li, Weiran Xu, Jun Guo, Cross-Document Coreference Resolution based on Automatic Text Summary, IEEE, 2010.
\vspace{0.1cm}

\textbf{[Kolya et al., 2010]} Anup Kumar Kolya, Asif Ekbal and Sivaji Bandyopadhyay, Event-Event Relation Identification: A CRF based Approach, IEEE, 2010.
\vspace{0.1cm}

\textbf{[Verhagen et al., 2007]} M.Verhagen, R.Gaizauskas, F.Schilder, M.Hepple, G.Katz and J.Pustejovsky, SemEval-2007 Task 15: TempEval Temporal Relation Identification, In Proceedings of the 4th International Workshop on Semantic Evaluations (SemEval-2007), pages 75-80, Prague, June, 2007, Association of Computational Linguistics.
\vspace{0.1cm}

\textbf{[Boguraev and Ando, 2005]} B.Boguraev and R.K.Ando, TimeMLCompliant Text Analysis for Temporal Reasoning, In Proceedings of Nineteenth International Joint Conference on Artificial Intelligence (IJCAI-05), pages 997-1003, Edinburgh, Scotland, August, 2005.
\vspace{0.1cm}

\textbf{[Mani et al., 2003]} I.Mani, B.Wellner, M.Verhagen, C.M.Lee, J.Pustejovsky, Machine Learning for Temporal Relation, In Proceedings of the 44th Annual meeting of the Association for Computational Linguistics, Sydney, Australia, 2003.
\vspace{0.1cm}

\textbf{[Chamber et al., 2007]} N.Chambers, S.Wang and D.Jurafsky,  Classifying Temporal Relations between Events, In Proceedings of the Association for Computational Linguistics Demo and Poster Sessions, pages 173-176, Prague, Czech Republic, June, 2007.
\vspace{0.1cm}

\textbf{[Huang et al., 2010]} Cuili Huang, Yaqiang Wang, Yongmei Zhang, Yu Jin and Zhongua Yu, Coreference Resolution in biomedical full-text articles with domain dependent features, 2nd International Conference on Computer Technology (ICCTD), 2010.
\vspace{0.1cm}

\textbf{[Shalom and Leass, 2004]} Lappin Salom and J.Herbert Leass, An algorithm for pronominal anaphora resolution, Computational Linguistics, 2004.
\vspace{0.1cm}

\textbf{[Kennedy and Boguraev, 1996]} Christopher Kennedy and Branimir Boguraev, Pronominal Anaphora Resolution without a Parser, In Proceedings of the 16th International Conference on Computational Linguistics, 1996.
\vspace{0.1cm}

\textbf{[Renata and Massimo, 2000]} Vieira Renata and Poesio Massimo, An Empirically based system for processing Definite Descriptions, Computational Linguistics, pages 525-579, 2000.
\vspace{0.1cm}

\textbf{[Caroline and Ted, 2008]} Gasperin Caroline and Briscoe Ted, Statistical anaphora resolution in biomedical texts, In Proceedings of the 22nd International Conference on Computational Linguistics (COLING), pages 257-264, Manchester, 2008.
\vspace{0.1cm}

\textbf{[Kong et al., 2010]} Fang Kong, Jianmei Zhou, Guodong Zhou and Qiaoming Zhu, Dependency tree based Anaphoricity determination for Coreference Resolution, International Conference on Asian Language Processing, 2010.
\vspace{0.1cm}

\textbf{[Etzioni et al., 2008]} O.Etzioni, M.Banko, S.Soderland and D.S.Weld, Open Information Extraction from the Web, ACM Communications, pages 68-74, 2008.
\vspace{0.1cm}

\textbf{[Feldman et al., 2008]} R.Feldman, Y.Regev and M.Gorodetsky, A modular Information Extraction System, Journal of Intelligent Data Analytics, pages 51-71, 2008.
\vspace{0.1cm}

\textbf{[Bodem et al., 2011]} Christoph Boden, Thomas Hafele and Alexander Loser, Classification Algorithms for Relation Prediction, ICDE Workshop, 2011.
\vspace{0.1cm}

\textbf{[Muludi et al., 2011]} Kurnia Muludi, Dwi H.Widyantoro, Kuspriyanto and Oerip S. Santoso, Multi-Inductive Learning Approach for Information Extraction, International Conference on Electrical Engineering and Informatics, Bandung, Indonesia, 17-19 July, 2011.
\vspace{0.1cm}

\textbf{[Amami et al., 2012]} Maha Amami, Aymen Elkhlifi and Rim Faiz, BioEv: A System for Learning Biological Event Extraction, International Conference on Information Technology and e-Services, 2012.
\vspace{0.1cm}

\textbf{[Burges, 1998]} C.J.C Burges, A tutorial on Support Vector Machines for Pattern Recognition, Data Mining and Knowledge Discovery, pages 121-167, 1998.
\vspace{0.1cm}

\textbf{[Lee et al., 2001]} T. Berners-Lee, J.Hendler and O.Lassila, The Semantic Web, In Scientific American, pp. 29-37, May, 2001.
\vspace{0.1cm}

\textbf{[Yang et al., 2015]} Bishan Yang, Claire Cardie and Peter Frazier, A Hierarchical Distance-dependent Bayesian Model for Event Coreference Resolution, Transactions of the Association for Computational Linguistics, Vol. 3, pp.517-528, 2015.
\vspace{0.1cm}

\textbf{[Bejan and Harabagiu, 2014]} Cosmin Adrian Bejan and Sanda Harabagiu, Unsupervised Event Coreference Resolution, Computational Linguistics, pp. 311-347, 2014.
\vspace{0.1cm}

\textbf{[Cardie and Wagstaff, 1999]} Claire Cardie and Kiri Wagstaff, Noun Phrase coreference as clustering, In Proceedings of the 1999 Joint SIGDAT Conference on Empirical Methods in Natural Language Processing and Very Large Corpora, pages 82-89, 1999.
\vspace{0.1cm}

\textbf{[Raghunathan et al., 2010]} Karthik Raghunathan, Heeyoung Lee, Sudarshan Rangarajan, Nathanael Chambers, Mihai Surdeanu, Dan Jurafsky, and Christopher Manning, A multi-pass sieve for coreference resolution, pages 492-501, In EMNLP, 2010.
\vspace{0.1cm}

\textbf{[Lee et al., 2011]} Heeyoung Lee, Yves Peirsman, Angel Chang, Nathanael Chambers, Mihai Surdeanu and Dan Jurafsky, Stanford's multi-pass sieve coreference resolution system at the CoNLL-2011 shared task, In Proceedings of the Fifteenth Conference on Computational Natural Language Learning: Shared Task, pages 28-34, 2011.
\vspace{0.1cm}

\textbf{[Stoyanov et al., 2009]} Veselin Stoyanov, Nathan Gilbert, Claire Cardie, and Ellen Riloff, Conundrums in noun phrase coreference resolution: Making sense of the state-of-the-art, In Proceedings of the Joint Conference of the 47th Annual Meeting of the ACL and the 4th International Joint Conference on Natural Language Processing of the AFNLP, Vol. 2, pages 656-664, 2009.
\vspace{0.1cm}

\textbf{[Rahman and Ng, 2011]} Altaf Rahman and Vincent Ng, Coreference resolution with world knowledge, In Association for Computational Linguistics, pages 814-824, 2011.
\vspace{0.1cm}

\textbf{[Durrett and Klein, 2010]} Greg Durrett and Dan Klein, Coreference resolution in a modular, entity-centered model, in NAACL, pages 385-393, 2010.
\vspace{0.1cm}

\textbf{[Haghighi and Klein, 2013]} Aria Haghighi and Dan Klein, Easy victories and uphill battles in coreference resolution, In EMNLP, pages 1971-1982, 2013.
\vspace{0.1cm}

\textbf{[Zelaia et al., 2015]} Ana Zelaia, Olatz Arregi and Basilio Sierra, A Multi-classifier Approach to support Coreference Resolution in a Vector Space Model, Proceedings of NAACL-HLT, 2015, pages 17-24, Denver, Colorado, Association for Computational Linguistics, 2015.
\vspace{0.1cm}

\textbf{[Orasan et al., 2008]} Constantin Orasan, Dan Cristea, Ruslan Mitkov and Antonio Branco, Anaphora Resolution Exercise: an Overview, Proceedings of the LREC, 2008.
\vspace{0.1cm}

\textbf{[Versley et al., 2008]} Yannick Versley, Simone Paolo Ponzetto, Massimo Poessio, Vladimir Eidelman, Alan Jern, Jason Smith, Xiaofeng Yang, and Alessandro Moschitti, Bart: a modular toolkit for coreference resolution, Proceedings of the HLT-Demonstrations, pp. 9-12, 2008.
\vspace{0.1cm}

\textbf{[Bengtson and Roth, 2008]} Eric Bengtson and Dan Roth, Understanding the value of features for coreference resolution, Proceedings of the EMNLP, pages. 294-303, 2008.
\vspace{0.1cm}

\textbf{[Manning et al., 2014]} Christopher D. Manning, Mihai Surdeanu, John Bauer, Jenny Finkel, Steven J. Bethard and David McClosky, The Stanford CoreNLP Natural Language Processing Toolkit, Proceedings of 52nd Annual Meeting of the Association for Computational Linguistics: System Demonstrations, pp. 55-60, 2014.
\vspace{0.1cm}

\textbf{[Soderland et al., 1995]} S.Soderland, D.Fisher, J.Aseltine, and W.Lehnert, CRYSTAL: Inducing a conceptual dictionary, In Proceedings of the Fourteenth International Joint Conference on Artificial Intelligence, pages 1314-21, 1995.
\vspace{0.1cm}

\textbf{[Califf and Mooney, 1997]} M.Califf and R.Mooney, Relational Learning of pattern-match rules for Information Extraction, In Workshop in Natural Language Learning, Association for Computational Linguistics, 1997.
\vspace{0.1cm}

\textbf{[Lafferty et al., 2001]} John D.Lafferty, Andrew McCallum, and Fernando C.N.Pereira, Conditional Random Fields: Probabilistic models for segmenting and labeling sequence data, In Proceedings of the Eighteenth International Conference on Machine Learning (ICML), 2001.
\vspace{0.1cm}

\textbf{[Mintz et al., 2009]} Mike Mintz, Steven Bills, Rion Snow, and Dan Jurafsky, Distant Supervision for relation extraction without labeled data, In Proceedings of Association for Computational Linguistics, 2009.
\vspace{0.1cm}

\textbf{[Bunescu and Mooney, 2007]} Razvan Bunescu and Raymond Mooney, Learning to extract relations from the web using minimal supervision, In Proceedings of the 45th Annual Meeting of the Association for Computational Linguistics, 2007.
\vspace{0.1cm}

\textbf{[Riedel et al., 2010]} Sebastian Riedel, Limin Yao, and Andrew McCallum, Modeling relations and their mentions without labeled text, In Proceedings of the Sixteenth European Conference on Machine Learning (ECML), pages 148-163, 2010.
\vspace{0.1cm}

\textbf{[Hoffman et al., 2011]} Raphael Hoffman, Congle Zhang, Xiao Ling, Luke Zettlemoyer, and Daniel S.Weld, Knowledge based weak supervision for information extraction of overlapping relations, In Proceedings of Association for Computational Linguistics, 2011.
\vspace{0.1cm}

\textbf{[Surdeanu et al., 2012]} Mihai Surdeanu, Julie Tibshirani, Ramesh Nallapati, and Christopher D.Manning, Multi-Instance Multi-Label learning for relation extraction, In Proceedings of EMNLP, pages 455-465, Association for Computational Linguistics, 2012.
\vspace{0.1cm}

\textbf{[Riedel et al., 2013]} Sebastian Riedel, Limin Yao, Andrew McCullum, and Benjamin M.Marlin, Relation extraction with matrix factorization and universal schemas, pages 74-84, NAACL-HLT, 2013.
\vspace{0.1cm}

\textbf{[Nguyen and Grisham, 2014]} Thien Huu Nguyen and Ralph Grisham, Employing word representations and regularization for domain adaptation of relation extraction, In Proceedings of the 52nd Annual Meeting of the Association for Computational Linguistics, volume 2, pages 68-74, 2014.
\vspace{0.1cm}

\textbf{[Angeli et al., 2014]} Gabor Angeli, Julie Tibshirani, Jean Y.Wu, and Christopher D.Manning, Combining distant and partial supervision for relation extraction, In EMNLP, 2014.
\vspace{0.1cm}

\textbf{[Zhang et al., 2012]} Ce Zhang, Feng Niu, Christopher Re, and Jude Shavlik, Big data versus the crowd: Looking for relationships in all the right places, In Proceedings of the 50th Annual Meeting of the Association for Computational Linguistics, Long papers, volume 1, pages 825-834, Association for Computational Linguistics, 2012.
\vspace{0.1cm}

\textbf{[Pershina et al., 2014]} Maria Pershina, Bonan Min, Wei Xu, and Ralph Grisham, Infusion of labeled data into distant supervision for relation extraction, In Proceedings of the Association for Computational Linguistics, 2014.
\vspace{0.1cm}

\textbf{[Alicia and Fahlman, 2007]} Alicia T. and Scott E. Fahlman, CMU-AT: Semantic Distance and Background Knowledge for Identifying Semantic Relations, Proceedings of the 4th International Workshop on Semantic Evaluations (SemEval-2007), pages 121-124, Prague, 2007.
\vspace{0.1cm}

\textbf{[Hendrickx et al., 2007]} I. Hendrickx, R.Morante, C.Sporleder, Antal V.D.Bosch, ILK: Machine Learning of semantic relations with shallow features and almost no data, Proceedings of the 4th International Workshop on Semantic Evaluations (SemEval-2007), pages 121-124, Prague, June, 2007.
\vspace{0.1cm}

\textbf{[Fellbaum, 1998]} C.Editor.Fellbaum, WordNet: An electronic lexical database and some of its applications, MIT Press, 1998.
\vspace{0.1cm}

\textbf{[Claudio et al., 2007]} G.Claudio, L.Alberto, P.Daniele, and R.Lorenza, FBK-IRST: Kernel Methods for Semantic Relation Extraction, Proceedings of the 4th International Workshop on Semantic Evaluations (SemEval-2007), pages 121-124, Prague, June, 2007.
\vspace{0.1cm}

\textbf{[Girju et al., 2005]} R.Girju, D.Moldovan, M.Tatu and D.Antohe, Automatic discovery of Part-Whole relations, Association for Computing Machinery, 2005.
\vspace{0.1cm}

\textbf{[Tesfaye et al., 2016]} Debela Tesfaye, Michael Zock, and Solomon Teferra, Combining syntactic patterns and Wikipedia's hierarchy of hyperlinks to extract meronym relations, Proceedings of NAACL-HLT 2016.
\vspace{0.1cm}

\textbf{[Hashimoto et al., 2015]} Kazuma Hashimoto, Pontus Stenetorp, Makoto Miwa, and Yoshimasa Tsuruoka, Task Oriented learning of word embeddings for semantic relation classification. In Proceedings of the Nineteenth Conference on Computational Natural Language Learning, pages 268-278, Beijing, China, ACL, 2015.
\vspace{0.1cm}

\textbf{[dos Santos et al., 2015]} Cicero dos Santos, Bing Xiang, and Bowen Zhou, Classifying relations by ranking with convolutional neural networks, In Proceedings of the 53rd Annual Meeting of the Association for Computational Linguistics and the 7th International Joint Conference on Natural Language Processing, volume 1, pages 626-634, ACL, 2015.
\vspace{0.1cm}

\textbf{[Socher et al., 2012]} Richard Socher, Bordy Huval, Christopher D.Manning and Andrew Y.Ng, Semantic compositionality through recursive matrix-vector spaces, In Proceedings of the 2012 Joint Conference on Empirical Methods in Natural Language Processing and Computational Natural Language Learning, pages 1201-1211, ACL, 2012.
\vspace{0.1cm}

\textbf{[Xu et al., 2015a]} Kun Xu, Yansong Feng, Songfang Huang, and Dongyan Zhao, Semantic relation classification via convolutional neural networks with simple negative sampling, In Proceedings of the 2015 Conference on Empirical Methods in Natural Language Processing, pages 536-540, ACL, 2015.
\vspace{0.1cm}

\textbf{[Xu et al., 2015b]} Yan Xu, Lili Mou, Ge Li, Yunchuan Chen, Hao Peng, and Zhi Jin, Classifying relations via LSTM networks along shortest dependency paths, In Proceedings of the 2015 Conference on Empirical Methods in Natural Language Processing, pages 1785-1794, ACL, 2015.
\vspace{0.1cm}

\textbf{[Tai et al., 2015]} Kai Sheng Tai, Richard Socher, and Christopher D.Manning, Improved semantic representations from tree-structured LSTM networks, In Proceedings of the 53rd Annual Meeting of the Association for Computational Linguistics, pages 1556-1566, Beijing, China, ACL, July, 2015.
\vspace{0.1cm}

\textbf{[Lin et al., 2016]} Yankai Lin, Shiqi Shen, Zhiyuan Liu, Huanbo Luan, and Maosong Sun, Neural Relation Extraction with Selective Attention over Instances,Proceedings of the 54th Annual Meeting of the Association for Computational Linguistics, pages 1105-1116, August 7-12, 2016.
\vspace{0.1cm}

\textbf{[Gamallo et al., 2012]} Pablo Gamallo, Marcos Garcia, and Santiago Fernandez-Lanza, Dependency based open information extraction, In Proceedings of ROBUS-UNSUP, 2012.
\vspace{0.1cm}

\textbf{[Tseng et al., 2014]} Yuan-Hsein Tseng, Lung-Hao Lee, Shu-Yen Lin, Boshun Liao, Mei-Jun Liu, Hsin-Hsi Chen, Oren Etzioni, and Anthony Fader, Chinese Open relation extraction for knowledge acquisition, In Proceedings of EACL, 2014.
\vspace{0.1cm}

\textbf{[Lewis and Steedman, 2013]} Mike Lewis and Mark Steedman, Unsupervised Induction of cross-lingual semantic relations, In Proceedings of EMNLP, 2013.
\vspace{0.1cm}

\textbf{[Gerber and Ngomo, 2012]} Daniel Gerber and Alex-Cyrille Ngonga Ngomo, Extracting multilingual natural language patterns for RDFs predicates, In Proceedings of the 18th International Conference on Knowledge Engineering and Knowledge Management, 2012.
\vspace{0.1cm}

\textbf{[Khirbat et al., 2016]} Gitansh Khirbat, Jianzhong Qi, and Rui Zhang, N-ary Biographical Relation Extraction using shortest path dependencies, In Proceedings of Australasian Language Technology Association Workshop, pages 74-83, 2016.
\vspace{0.1cm}

\textbf{[McDonald et al., 2005]} Ryan McDonald, Fernando Pereira, Seth Kulick, Scott Winters, Yang Jin, and Pete White, Simple algorithms for complex relation extraction with applications to bio-medical information extraction, In Proceedings of the 43rd Annual Meeting of the Associations for the Computational Linguistics, 2005.
\vspace{0.1cm}

\textbf{[Li et al., 2015]} Hong Li, Sebastian Krause, Feiyu Xu, Andrea Moro, Hans Uszkoreit ,and Roberto Navigli, Improvement of n-ary relation extraction by adding lexical semantics to distant supervision rule learning, In Proceedings of the 7th International Conference on Agents and Artificial Intelligence, 2015.
\vspace{0.1cm}

\textbf{[Li et al., 2016]} Zhuang Li, Lizhen Qu, Qiangkai Xu, and Mark Johnson, Unsupervised pre-training with Seq2Seq reconstruction loss for deep relation extraction models,  In Proceedings of Australasian Language Technology Association Workshop, pages 54-64, 2016.
\vspace{0.1cm}

\textbf{[Shen and Huang, 2016]} Yatian Shen and Xuanjing Huang, Attention based Convolutional Neural Network for Semantic Relation Extraction, Proceedings of the COLING, 26th Conference on Computational Linguistics, Technical papers, pages 2526-2536, December, 2016.
\vspace{0.1cm}

\textbf{[Mirza and Tonelli, 2016]} Paramita Mirza and Sara Tonelli, CATENA: CAusal and TEmporal relation extraction from NAtural language texts, Proceedings of the COLING,26th Conference on Computational Linguistics, Technical papers, pages 64-75, December, 2016. 
\vspace{0.1cm}

\textbf{[D'Souza and Ng, 2013]} Jennifer D'Souza and Vincent Ng, Classifying temporal relations with rich linguistic knowledge, In Proceedings of the 2013 Conference of the North American Chapter of the Association for Computational Linguistics: Human Language Technologies, pages 918-927, ACL, 2013.
\vspace{0.1cm}

\textbf{[Chamber et al., 2014]} Nathanael Chambers, Taylor Cassidy, Bill McDowell, and Steven Bethard, Dense event ordering with a multi-pass architecture, Transactions of the Association for Computational Linguistics, 2014.
\vspace{0.1cm}

\textbf{[Bollacker et al. 2008]} Kurt Bollacker, Colin Evans, Praveen Paritosh, Tim Sturge, and Jamie Taylor. 2008. Freebase: A Collaboratively Created Graph Database for Structuring Human Knowledge. In Proceedings of the 2008 ACM SIGMOD International Conference on Management of Data, pages 1247-1250.
\vspace{0.1cm}

\textbf{[Bordes et al. 2011]} Antoine Bordes, Jason Weston, Rnan Collobert, and Yoshua Bengio. 2011. Learning Structured Embeddings of Knowledge Bases. In Proceedings of the Twenty-Fifth AAAI Conference on Artificial Intelligence, pages 301-306.
\vspace{0.1cm}

\textbf{[Bordes et al. 2012]} Anotoine Bordes, Xavier Glorat, Jason Weston, and Yoshua Bengio. 2012. A Semantic Matching Energy Function for Learning with Multi-relational Data. Machine
Learning, 94(2): 233-259.
\vspace{0.1cm}

\textbf{[Bordes et al. 2013]} Antoine Bordes, Nicolas Usunier, Alberto Garcia-Duran, Jason Weston, and Oksana Yakhnenko. 2013. Translating Embeddings for Modeling Multi-relational Data. In Advances in Neural Information Processing Systems 26, pages 2787-2795.
\vspace{0.1cm}

\textbf{[Carlson et al. 2010]} Andrew Carlson, Justin Betteridge, Bryan Kisiel, Burr Settles, Estevam R Hruschka, Jr, and Tom M.Mitchell. 2010. Towards an Architecture for Never –Ending Language Learning. In Proceedings of the Twenty-Fourth AAAI Conference on Artificial Intelligence, pages 1306-1313.
\vspace{0.1cm}

\textbf{[Guu et al. 2015]} Kelvin Guu, John Miller, and Percy Liang. 2015. Traversing Knowledge Graphs in Vector Space. In Proceedings of the 2015 Conference on Empirical Methods in Natural Language Processing, pages 318-327.
\vspace{0.1cm}

\textbf{[Miller 1995]} George A Miller. 2015. WordNet: A Lexical Database for English. Communications of the ACM, 38(11):39-41
\vspace{0.1cm}

\textbf{[Mikolov et al. 2013]} Tomas Mikolov, Wen-tau Yih, and Geoffrey Zweig. 2013. Linguistic Regularities in Continuous Space Word Representations. In Proceedings of the 2013 Conference of the North American Chapter of the Association for Computational Linguistics: Human Language Technologies, pages 746-751.
\vspace{0.1cm}

\textbf{[Yang et al. 2015]} Bishan Yang, Wen-tau Yih, Xiaodong He, Jianfeng Gao, and Li Deng. 2015. Embedding Entities and Relations for Learning and Inference in Knowledge Bases. In Proceedings of the International Conference on Learning Representations.
\vspace{0.1cm}

\textbf{[Wang et al. 2014]} Zhen Wang, Jianwen Zhang, Jianlin Feng, and Zheng Chen. 2014. Knowledge Graph Embedding by Translating on Hyperplanes. In Proceedings of the Twenty-Eighth AAAI Conference on Artificial Intelligence, pages 1112-1119.
\vspace{0.1cm}

\textbf{[Toutanova et al. 2016]} Kristina Toutanova, Xi Victoria Lin, Wen-tau Yih, Hoifung Poon, and Chris Quirk. 2016. Compositional Learning of Embeddings for Relation Paths in Knowledge Bases and Text. In Proceedings of the 54 th Annual Meeting of the Association for Computational
Linguistics, June. 2016.
\vspace{0.1cm}

\textbf{[Socher et al. 2013]} Richar Socher, Danqi Chen, Christopher D Manning, and Andrew Ng. 2013. Reasoning with Neural Tensor Networks for Knowledge Base Completion. In Advances in Neural Information Processing Systems, pages 926-934.
\vspace{0.1cm}

\textbf{[Suchanek et al. 2007]} Fabian M Suchanek, Gjergji Kasneci, and Gerherd Weikum. 2007. YAGO: A Core of Semantic Knowledge. In Proceedings of the 16 th International Conference on World Wide Web, pages 697-706.
\vspace{0.1cm}

\textbf{[Nickel et al. 2011]} Maxmilian Nickel, Volker Tresp, and Hans-Peter-Kriegel. 2011. A Three- Way Model for Collective Learning on Multi-Relational Data. In Proceedings of the 28 th International Conference on Machine Learning, pages 809-816.
\vspace{0.1cm}

\textbf{[Nickel et al. 2015]} Maxmilian Nickel, Kevin Murphy, Volker Tresp, and Evgeniy Gabrilovich. 2015. A Review of Relational Machine Learning for Knowledge Graphs. Proceedings of the IEEE, 2016.
\vspace{0.1cm}

\textbf{[Chang et al. 2014]} Kai-Wei Chang, Wen-tau Yih, Bishan Yang, and Christopher Meek. 2014. Typed Tensor Decomposition of Knowledge Bases for Relation Extraction. In Proceedings of the 2014 Conference on Empirical Methods in Natural Language Processing(EMNLP), pages 1568-1579.
\vspace{0.1cm}

\textbf{[Nguyen et al. 2016]} Dat Quoc Nguyen, Kairit sirts, Lizhen Qu, and Mark Johnson. 2016. StransE: a novel embedding model of entities and relationships in knowledge bases. In Proceedings of the 15 th Annual Conference of the North American Chapter of the Association for Computational Linguistics: Human Language Technologies.
\vspace{0.1cm}

\textbf{[Lin et al. 2015a]} Yankai Lin, Zhiyuan Liu, Huanbo Luan, Maosong Sun, Siwei Rao, and Song Liu. 2015a. Modeling Relation Paths for Representation Learning of Knowledge Bases. In Proceedings of the 2015 Conference on Empirical Methods in Natural Language Processing,
pages 705-714.
\vspace{0.1cm}

\textbf{[Nguyen et al. 2016]} Dat Quoc Nguyen, Kairit Sirts, Lizhen Qu, and Mark Johnson. 2016. Neighborhood Mixture Model for Knowledge Base Completion. In Proceedings of the 2016 Conference on Natural Language Learning (CoNLL), 2016.
\vspace{0.1cm}

\textbf{[Xiao et al. 2016]} Han Xiao, Minlie Huang, and Xiaoyan Zhu. 2016. TransG: A Generative Model for Knowledge Graph Embedding. In Proceedings of the 54 th Annual Meeting of the Association for Computational Linguistics, pages 2316-2325.
\vspace{0.1cm}

\textbf{[Lin et al. 2015]} Yankai Lin, Zhiyuan Liu, Maosong Sun, Yang Liu, and Xuan Zhu. 2015. Learning Entity and Relation Embeddings for Knowledge Graph Completion. In Proceedings of the Association of the Advancement of Artificial Intelligence, 2015.
\vspace{0.1cm}

\textbf{[Bolla M, 1993]} M. Bolla. Spectra, Euclidean representations and clusterings of hypergraphs. Dis-
crete Mathematics, 117(1-3), 1993.
\vspace{0.1cm}

\textbf{[Zien et al., 1999]} J. Y. Zien, M. D. F. Schlag, and P. K. Chan. Multi-level spectral hypergraph
partitioning with arbitrary vertex sizes. IEEE Transactions on Computer-Aided Design of Integrated Circuits and Systems, 18(9):1389–1399, 1999.
\vspace{0.1cm}

\textbf{[Wimalasuriya and Dou, 2010]} C.Daya Wimalasuriya and Dejing Dou, Ontology based information extraction: An Introduction and a survey of current approaches, Journal of Information Science, June 2010, 36:306-323
\vspace{0.1cm}

\textbf{[Assal et al., 2011]} Hisham Assal, John Seng, Franz Kurfess, Emily Schwarz, and Kym Pohl, Semantically-Enhanced Information Extraction, IEEE, 2011.
\vspace{0.1cm}

\textbf{[Miwa and Bansal, 2016]} Makoto Miwa and Mohit Bansal, End-to-End Relation Extraction using LSTMs on Sequences and Tree Structures, Proceedings of the 54th Annual Meeting of the Association for Computational Linguistics, pages 1105-1116, August 7-12, 2016.
\end{document}